%% file: main.tex
\documentclass[10pt,twocolumn,letterpaper]{article}

\usepackage{iccv}              % To produce the CAMERA-READY version

\definecolor{iccvblue}{rgb}{0.21,0.49,0.74}
\usepackage[pagebackref,breaklinks,colorlinks,allcolors=iccvblue]{hyperref}

\usepackage{comment}
\usepackage{algorithm}
\usepackage{algpseudocode}
\usepackage{adjustbox}

\def\paperID{1736} % *** Enter the Paper ID here
\def\confName{ICCV}
\def\confYear{2025}

\title{LieHMR: Autoregressive Human Mesh Recovery with $SO(3)$ Diffusion}

\author{Donghwan Kim\;\;\;\;\; Tae-Kyun Kim\\
School of Computing, KAIST\\
\text{\{donghwan.kim, kimtaekyun\}@kaist.ac.kr}
}

\begin{document}

\maketitle
\input{sec/0_abstract}    
\input{sec/1_introduction}
\input{sec/2_related_work}

\input{sec/3_preliminary}
\input{sec/4_method}

\input{sec/5_experiments}
\input{sec/6_conclusion}
{
    \small
    \bibliographystyle{ieeenat_fullname}
    \bibliography{main}
}

% % WARNING: do not forget to delete the supplementary pages from your submission 
\input{sec/X_suppl}

\end{document}

%% file: sec/0_abstract.tex
\begin{abstract}
    We tackle the problem of Human Mesh Recovery (HMR) from a single RGB image, formulating it as an image-conditioned human pose and shape generation. While recovering 3D human pose from 2D observations is inherently ambiguous, most existing approaches have regressed a single deterministic output. Probabilistic methods attempt to address this by generating multiple plausible outputs to model the ambiguity. However, these methods often exhibit a trade-off between accuracy and sample diversity, and their single predictions are not competitive with state-of-the-art deterministic models. To overcome these limitations, we propose a novel approach that models well-aligned distribution to 2D observations. In particular, we introduce $SO(3)$ diffusion model, which generates the distribution of pose parameters represented as 3D rotations unconditional and conditional to image observations via conditioning dropout. Our model learns the hierarchical structure of human body joints using the transformer. Instead of using transformer as a denoising model, the time-independent transformer extracts latent vectors for the joints and a small MLP-based denoising model learns the per-joint distribution conditioned on the latent vector. We experimentally demonstrate and analyze that our model predicts accurate pose probability distribution effectively.
\end{abstract}

%% file: sec/1_introduction.tex
\section{Introduction}
\label{sec:intro}

Capturing and generating human motion is a fundamental task in computer vision with applications to metaverse, video-game modeling, and understanding daily actions. Recently, there are lots of works from pose estimation \cite{baek2020, armagan2020ECCV, 2024graspnet} to detailed reconstruction with texture maps \cite{kim2024bitt}, implicit functions \cite{lee2023im2hands, lee2023fourierhandflow, kim2025srhand}, or gaussian splatting \cite{lee2025MPMAvatar}. However, estimating 3D human pose and shape from monocular images is still an ill-posed problem due to depth ambiguity, occlusion, and truncation. Many approaches adopt statistical body models \cite{SMPL:2015, MANO:SIGGRAPHASIA:2017, SMPL-X:2019, ghum}, enabling neural networks to reconstruct 3D human meshes by predicting pose and shape parameters. A key challenge in these parametric methods is learning 3D rotation representations (\textit{e.g.}, axis-angle \cite{kanazawaHMR18}, rotation matrices \cite{levinson20neurips}, 6D vectors \cite{zhou2019cvpr, kolotouros2019spin, li2022cliff}) \cite{kolotouros2019cmr, Choi_2020_ECCV_Pose2Mesh, he24nrdf}. Alternatively, non-parametric methods directly predict the 3D coordinates of mesh vertices instead of the 3D rotation parameters \cite{kolotouros2019cmr, Moon_2020_ECCV_I2L-MeshNet, lin2021end-to-end, lin2021-mesh-graphormer, cho2022FastMETRO, corona2022learned, Ma_2023_CVPR}.

Despite significant progress in HMR, the single-output of deterministic methods cannot capture the inherent ambiguity due to the loss of depth information and occlusions in 2D images. To address this, several parametric approaches model uncertainty by learning probability distributions over body parameters and sample multiple plausible 3D outputs, which is useful in downstream tasks \cite{sengupta2021probabilistic, mhcdiff2024}. These learned distributions improve performance in challenging cases (\textit{e.g.}, heavy occlusions, egocentric views). However, they require multiple samples and aggregation to achieve state-of-the-art performance on large datasets compared to single-output HMR models.

Inspired by the success of autoregressive models in natural language processing \cite{Radford2018gpt, Radford2019gpt2, brown2020languagemodelsfewshotlearners}, vector quantization (VQ) \cite{oord2018neuraldiscreterepresentationlearning, razavi2019generatingdiversehighfidelityimages} has been adopted to discretize continuous data. Recent works apply tokenization to pose and vertex regression, transforming them into token prediction tasks \cite{dwivedi_cvpr2024_tokenhmr, fiche2024vq, fiche2024mega}. The vector quantized-variational autoencoder (VQ-VAE) learns a "codebook" from large-scale motion capture datasets in self-supervised manner, encoding valid pose priors.  While achieving state-of-the-art performance, these methods face an inherent trade-off between information loss and codebook complexity.

\begin{figure}
  \centering
  \includegraphics[width=1.0\linewidth]{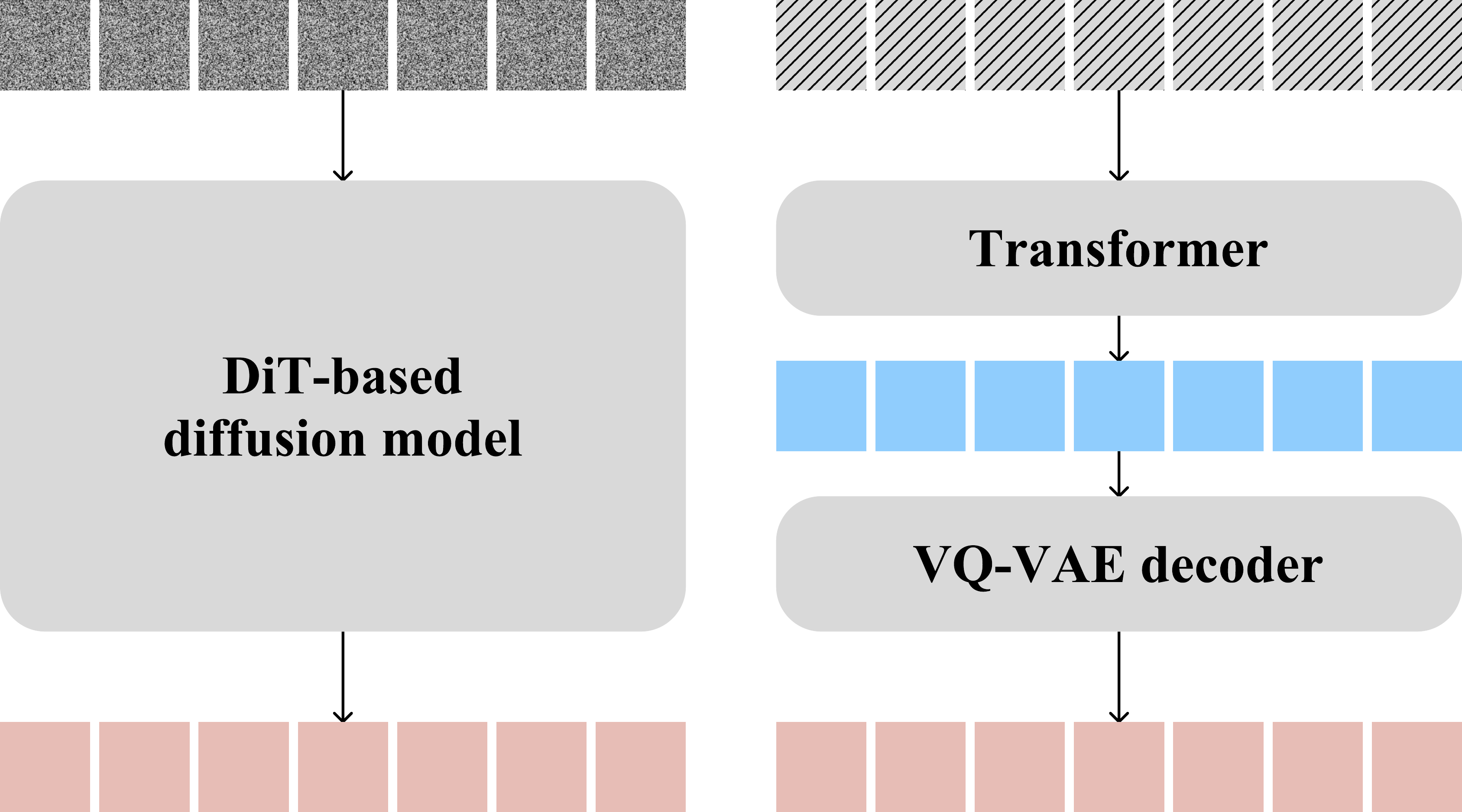}
  \caption{
    \textbf{(Left)} We modify LieHMR based on DiT \cite{Peebles2022DiT}. Here, the denoising model consists of the transformer and learns the joint probability distribution of whole tokens. We concatenate the image features to the input sequence. \textbf{(Right)} This is a brief overview for Vector-Quantization based methods \cite{dwivedi_cvpr2024_tokenhmr, fiche2024vq, fiche2024mega}. They predicts the quantized tokens conditioned on the image features from a fully masked sequence and reconstruct the pose parameters or 3D mesh with VQ-VAE decoder.
  }
  \label{fig:models}
\end{figure}

To address the aforementioned limitations, we propose LieHMR. (1) We utilize the denoising diffusion probabilistic model \cite{ho2020denoising} on the $SO(3)$ manifold, enabling robust 3D rotation learning and generating well-aligned pose distributions. The predicted distribution is narrow, resembling a Dirac delta, for less ambiguous images and broader to capture multiple plausible solutions for more ambiguous images. As a result, LieHMR achieves strong performance over both deterministic and probabilistic methods without requiring intermediate representations or model modifications. (2) We disentangle the transformer \cite{attention} and denoising model. Unlike transformer-based diffusion approaches \cite{bao2022all, Peebles2022DiT} in \cref{fig:models} (Left), our transformer is time-independent and captures joint relationships, while our denoising model is time-dependent and learns per-joint probability distributions. This disentanglement improves inference efficiency while effectively modeling joint distributions.

\begin{figure}
    \centering
    \includegraphics[width=1.0\linewidth]{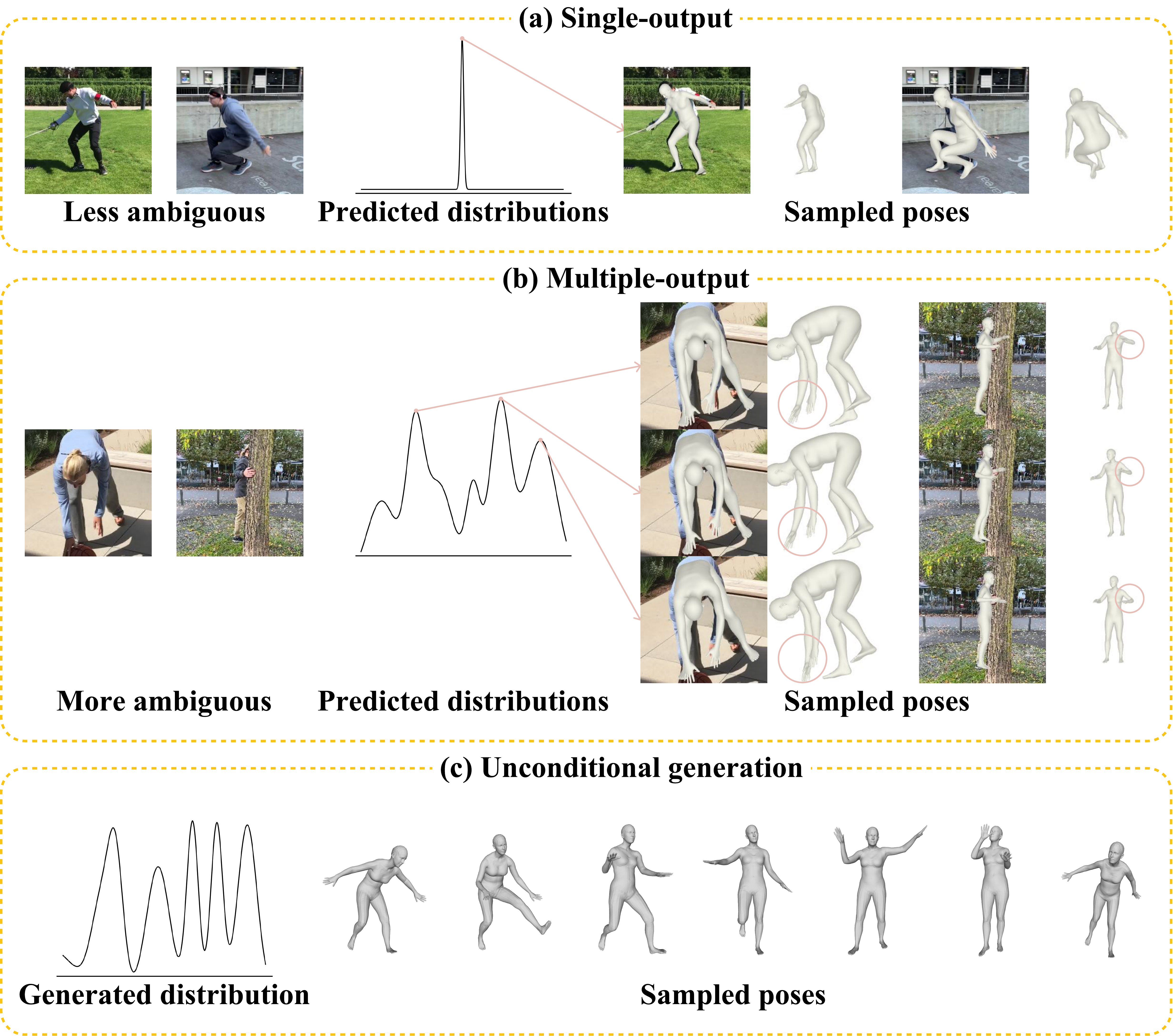}
    \caption{\textbf{Well-aligned probability distribution modeling to 2D observations in HMR.} We propose LieHMR, which simultaneously learns image-conditioned and unconditional human pose and shape generation. Our goal is modeling well-aligned distribution to 2D observations. (a) Given an image with less ambiguity, we can sample as accurate single output as prior deterministic HMR methods. (b) Given an ambiguous image, we can sample multiple plausible outputs. The diversity is mainly related to the depth ambiguity or occlusion. (c) LieHMR can also generate human pose and shape in unconditional manner, which states that the model learns the pose priors well.}
    \label{fig:teaser}
\end{figure}

We train LieHMR on standard HMR datasets in supervised manner and large-scale motion capture datasets in self-supervised manner, simultaneously.\footnote{In this paper, we consider learning conditional generation as supervised training and unconditional generation as self-supervised training.} Following MEGA \cite{fiche2024mega}, we perform extensive experiments to evaluate the image-conditioned and unconditional human pose and shape generation. Our experiments on in-the-wild HMR benchmarks demonstrate that LieHMR models well-aligned distribution to 2D observations. For unconditional generation, LieHMR shows comparable realism and diversity compared to previous methods learning human pose priors. We also analyze how LieHMR learns the probability distribution unconditional and conditional to image observations.

Our main contributions are as follows:

\begin{itemize}
    \item We introduce LieHMR, a novel generative framework for the image-conditioned and unconditional human pose and shape generation. It models well-aligned distribution and outperforms on single-output/multiple-output HMR and unconditional pose generation.

    \item LieHMR is based on the denoising diffusion probabilistic model on the $SO(3)$ manifold, which learns 3D rotation representations well.

    \item Our model consists of the transformer to learn joint relationships and the denoising model to learn per-joint probability distributions.
\end{itemize}

%% file: sec/2_related_work.tex
\begin{figure*}
  \centering
  \includegraphics[width=0.95\linewidth]{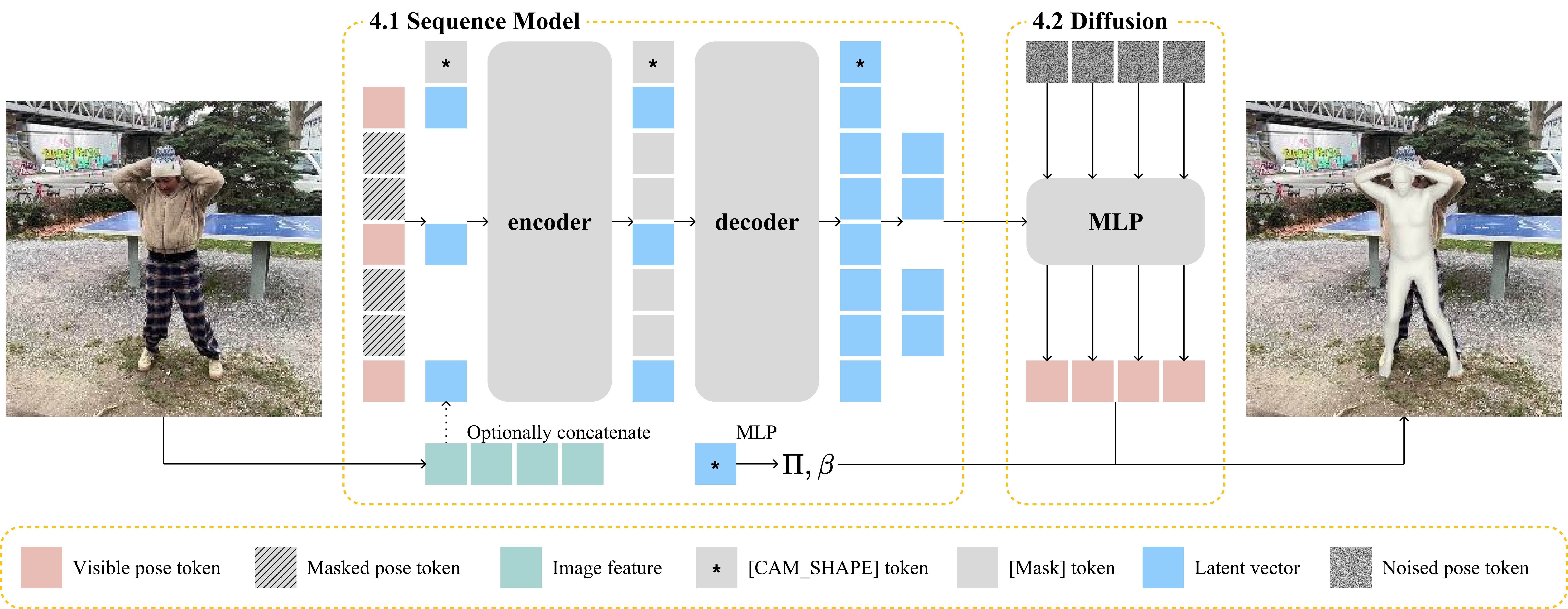}
  \caption{
    \textbf{Overview of LieHMR.} (\cref{subsec:sequence}) Given the partially visible pose tokens, the transformer-based sequence model extracts latent vectors $z$ for all tokens. Following the pipeline of Masked Autoencoder (MAE), we apply the encoder on the visible tokens and fill the mask tokens before the decoder. (\cref{subsec:so3_diffusion}) Given the noised pose token $\theta_t^i$, the MLP-based denoising model predicts the noise conditioned on the latent vector $z^i$. We perform the denoising process independently for each token. The forward and reverse process of diffusion procedure are constrained on $SO(3)$ manifold. We optionally concatenate the image features for image-conditioned generation.
  }
  \label{fig:architecture}
\end{figure*}

\section{Related Work}
\subsection{Regression-based methods}

Following the HMR \cite{kanazawaHMR18} model, most learning-based approaches reconstruct 3D human meshes from images using neural networks in regression manner. Parametric approaches predict the pose and shape parameters of the SMPL model. Several works aim to reduce the gaps between the image and 3D rotation representations, introducing intermediate representations, such as 3D keypoints \cite{madadi2019smplrdeepsmplreverse, iqbal2021kama, li2021hybrik}, 2D heatmaps \cite{pavlakos2018learningestimate3dhuman}, or part segmentation \cite{Kocabas_PARE_2021}. Instead of 3D joint rotations, non-parametric approaches predict vertices to
reconstruct 3D humans. They leverage graph convolutional neural networks \cite{kolotouros2019cmr, Moon_2020_ECCV_I2L-MeshNet, lin2021-mesh-graphormer} or transformers \cite{lin2021end-to-end, cho2022FastMETRO, Ma_2023_CVPR} to learn the topology of human meshes. However, they show the lack of generalization on the out-of-distribution \cite{lin2021end-to-end, fiche2024vq}.

To address the uncertainty from depth ambiguities and occlusions, several works generate the probability distribution. They employ the generative models, such as conditional variational autoencoder \cite{Sharma_2019_ICCV}, normalizing flows \cite{kolotouros2021prohmr, dwivedi_3dv2023_poco}, and diffusion models \cite{zhang2023probabilistic, Cho_2023_ICCV, xu2024scorehypo, stathopoulos2024score, lee2025rewind}. Some probabilistic approaches also constrain on $SO(3)$ manifold with the matrix Fisher distribution \cite{sengupta2021hierprobhuman, fang2023propose} or in normalizing flows \cite{sengupta2023humaniflow}. These methods can predict diverse plausible solutions from ambiguous 2D observations well. However, their outputs from clear 2D observations are also too diverse, struggling on large datasets when sampling a single output.

\subsection{Vector-Quantization based methods}
Several works in human motion generation \cite{yang2023QPGesture, zhang2023generating, jiang2024motiongpt}, following autoregressive generation in large language models, use VQ-VAE \cite{oord2018neuraldiscreterepresentationlearning} for quantizing human motion. Recent works in HMR also adopt VQ-VAE to discretize the pose parameters or 3D meshes. TokenHMR \cite{dwivedi_cvpr2024_tokenhmr} quantizes the pose parameters and learns the valid pose priors. VQ-HPS \cite{fiche2024vq} quantizes 3D meshes and trains the model using only cross-entropy loss. MEGA \cite{fiche2024mega}, which is concurrent work to ours, quantizes 3D meshes and trains the model as a Masked Autoencoder (MAE) \cite{MaskedAutoencoders2021}. However, the evaluation for unconditional generation includes the shape parameter, which is less important for human motion generation, and they need different inference procedure to achieve the performance. Their training pipeline has three stages: training VQ-VAE, self-supervised pre-training, and supervised training. They usually transformers to predict tokens and VQ-VAE decoder reconstructs the pose parameters or 3D meshes from predicted tokens (in \cref{fig:models} right).

\subsection{Pose priors}
Generative approaches learn the probability distribution to model the valid space of human poses. Several works use Gaussian Mixture Models (GMM) \cite{bogo2016smplautomaticestimation3d}, Generative Adversarial Networks (GAN) \cite{georgakis2020hierarchicalkinematichumanmesh, kanazawaHMR18}, VAE \cite{SMPL-X:2019}, and normalizing flows \cite{kolotouros2021prohmr} as priors to constrain the valid space during training. HuProSO’3 \cite{dünkel2024normalizingflowsproductspace} introduces a normalizing flow on SO(3) manifold to learn the per-joint probability distribution. Pose-NDF \cite{tiwari22posendf} and NRDF \cite{he24nrdf} learn the space of plausible poses as the zero-level-set of a neural implicit function. GFPose \cite{ci2022gfpose} and DPoser \cite{lu2023dposer} are diffusion-based methods learning the pose priors for single human, while BUDDI \cite{mueller2023buddi} and InterHandGen \cite{lee2024interhandgen} introduce diffusion models to learn human interaction and hand interaction, respectively.

%% file: sec/3_preliminary.tex
\section{Preliminary}

\paragraph{Diffusion models on $SO(3)$ group.}
Over the past years, denoising diffusion probabilistic models (DDPMs) \cite{ho2020denoising} have been applied on non-Euclidean space \cite{leach2022denoising, jagvaral2023diffusion, urain2022se3dif, yim2023se, jiang2023se, hsiao2024confronting}. Hsiao et al. \cite{hsiao2024confronting} propose a score-based diffusion method on $SO(3)$/$SE(3)$ group, which is the first application to the image domain. The main differences compared to diffusion models on Euclidean space are forward and reverse process:

\begin{equation}
\label{Eq:so3_forward}
x_t = x_0 \text{Exp}(\sqrt{\alpha_t} \text{Log}(\epsilon)),
\end{equation}

\begin{equation}
\label{Eq:so3_reverse}
\begin{split}
\hat{x}_{t-1} =
\underbrace{x_t (\text{Exp}(\sqrt{\alpha_t} \text{Log}(\hat{\epsilon})))^{-1}}_{\text{predicted }x_0}
&\underbrace{\text{Exp}(\sqrt{\alpha_{t-1} - \sigma_t^2} \text{Log}(\hat{\epsilon}))}_{\text{direction pointing to }x_t}\\
&\underbrace{\text{Exp}(\sigma_t \text{Log}(\epsilon))}_{\text{random noise}}.
\end{split}
\end{equation}

Here, $x_0$ and $x_t$ are original and noised data, respectively, and $\alpha_t$ is pre-defined noise schedule like original diffusion process. $\hat{\epsilon}$ is predicted noise and $\sigma_t = \eta * \sqrt{\frac{\alpha_{t-1}  (\alpha_t - \alpha_{t-1})}{\alpha_t}}$ is from DDIM sampling \cite{song2022denoising}. There is a one-to-one correspondence between Lie group $SO(3)$ and Lie algebra (tangent space) $\mathfrak{so}(3)$: the logarithm map $\text{Log}: SO(3) \rightarrow \mathfrak{so}(3)$ and the exponential map $\text{Exp}: \mathfrak{so}(3) \rightarrow SO(3)$. The diffusion procedure scales the noise on Lie algebra (linear vector space), not Lie group (non-linear manifold space). 

\paragraph{Autoregressive models without Vector Quantization \cite{li2024autoregressive}.}
Autoregressive models in continuous-valued domains have intensely focused on discretizing the data through vector quantization (VQ) \cite{oord2018neuraldiscreterepresentationlearning, razavi2019generatingdiversehighfidelityimages}. However, MAR \cite{li2024autoregressive} observes that autoregressive models need per-token probability distribution. Specifically, the models are trained with a loss function (\textit{e.g.}, cross-entropy loss) which measures the distribution, and sample data from the generated distribution (\textit{e.g.}, softmax with a temperature). MAR \cite{li2024autoregressive} proposes to model the per-token probability distribution in continuous-valued space by diffusion procedure \cite{ho2020denoising}. Specifically, MAR can be divided to two components: (1) a transformer \cite{attention} predicts latent vector $z_i$ conditioned on previous tokens $x_{<i}$, and (2) a small denoising model generates the next token $x_i$ conditioned on the latent vector.

%% file: sec/4_method.tex
\section{Method}

We aim at generating human pose $\theta = [\theta^1,...,\theta^{24}] \in \mathbb{R}^{24 \times 3 \times 3}$ and shape parameter $\beta \in \mathbb{R}^{10}$, including unconditional generation and image-conditioned generation. Since the ambiguity and diversity of human mesh is mainly related to pose, we adopt diffusion model on SO(3) group to learn the probability distribution of pose parameters, which are represented as 3D rotations. LieHMR consists of a transformer-based sequence model, which learns the hierarchical relationship of human body joints, and a small MLP-based denoising model, which learns the per-joint probability distribution, as illustrated in \cref{fig:architecture}.

\subsection{Sequence model}
\label{subsec:sequence}
The sequence model follows the overall pipeline of standard Masked Autoencoder (MAE). We apply an MAE-style encoder on the sequence of the visible pose tokens $\theta_{vis} \in \mathbb{R}^{N \times d}$, a token for camera and shape parameters $[\texttt{CAM\_SHAPE}] \in \mathbb{R}^d$, and the image features $g(I) \in \mathbb{R}^{WH \times d}$ (with positional embedding). Then, we fill mask tokens $[\texttt{MASK}] \in \mathbb{R}^d$ on the masked position of the encoded sequence and map this sequence with an MAE-style decoder. We predict the camera parameter $\pi$ and shape parameter $\beta$ from the decoded $[\texttt{CAM\_SHAPE}]$ with an MLP-based regressor. The decoded pose tokens (with positional embedding added again) serve as  conditioning latent vectors $z=[z^1,...,z^J] \in \mathbb{R}^{J \times d}$ for the denoising model. The attention mechanism allows all tokens to see the visible pose tokens and image features, so the latent vectors capture the relationship of human body joints, image observation, and which joint to predict by positional embeddings. Therefore, the sequence model generates the latent vector, camera parameter and shape parameter, given the known pose tokens and image features:

\begin{equation}
\label{Eq:autoregressive}
z^i,\pi,\beta=f(\theta^{1},...,\theta^{i-1}|g(I)).
\end{equation}

\subsection{$SO(3)$ diffusion}
\label{subsec:so3_diffusion}
Conditioned on the latent vector $z^i \in \mathbb{R}^d$ from the sequence model, the diffusion model $\epsilon_\phi$ learns the per-joint probability distribution by denoising the pose parameter:

\begin{equation}
\label{Eq:so3_diffusion}
\epsilon_\phi(\theta^i_t, z^i, t)=\hat{\epsilon},
\end{equation}
where $\theta^i_t$ is the noised pose parameter of $i$-th joint, $t$ is diffusion step, and $\hat{\epsilon}$ is predicted noise. The forward and reverse process of diffusion procedure are constrained on $SO(3)$ manifold (\cref{Eq:so3_forward} and \cref{Eq:so3_reverse}).

\subsection{Training strategy}
LieHMR is trained in a Masked Autoencoder (MAE) manner with the diffusion loss. The original pose tokens $\theta$ are randomly masked and the sequence model extracts the latent vectors $z$, camera parameter $\pi$, and shape parameter $\beta$. We noise the original pose tokens $\theta$ with random noise $\epsilon$ and diffusion step $t$ by \cref{Eq:so3_forward} and the denoising model predicts the noise $\hat{\epsilon}$. We can obtain the predicted joints $\hat{J}_{3D}$ from $\beta$ and predicted $\hat{\theta}_0$ by \cref{Eq:so3_reverse} and project the joints to $\hat{J}_{2D}$ with $\pi$. The total objective consists of the diffusion loss, 3D joints loss, and 2D joints loss:\footnote{We set  $\lambda_{\text{diff}}=1, \lambda_{\text{3D}}=1, \text{and } \lambda_{\text{2D}}=1$.}

\begin{equation}
\label{Eq:objective}
\begin{split}
\mathcal{L}= \lambda_{\text{diff}}\underbrace{\|\epsilon-\hat{\epsilon}\|_2^2}_{\text{diffusion loss}} &+ \lambda_{\text{3D}}\underbrace{\|J_{3D}-\hat{J}_{3D}\|_1}_{\text{3D joints loss}} \\ &+ \lambda_{\text{2D}}\underbrace{\|J_{2D}-\hat{J}_{2D}\|_1}_{\text{2D joints loss}}.
\end{split}
\end{equation}

In addition to supervised learning for image-conditioned generation, we train the model in a self-supervised manner for unconditional generation using an extensive mocap dataset. For unconditional generation, we set $\lambda_{\text{2D}}=0$ since there are no corresponding images for the pose and shape parameters. We only concatenate the image features $g(I)$ to the input sequence for the supervised training samples. This training strategy can be considered as conditioning dropout \cite{ho2022classifierfreediffusionguidance}, resulting in a unified model that learns both with and without conditioning in single training stage.

\subsection{Generation strategy}
LieHMR generates human pose and shape parameters in an autoregressive manner. We start from a fully masked sequence of pose tokens and gradually generate a sequence of tokens. We can formulate the joint probability distribution of the parameters:

\begin{equation}
\label{Eq:probability}
p(\theta^1,...,\theta^J,\pi,\beta)=\prod_{i=1}^{J}p(\theta^i,\pi,\beta|\theta^1,....,\theta^{i-1},g(I)).
\end{equation}

Instead of generating one token at a time, we predicts multiple tokens simultaneously in random order, which is called masked autoregressive (MAR) \cite{li2024autoregressive}. We model the per-joint probability distribution with diffusion procedure, controlling the diversity with $\eta$ of DDIM sampling \cite{song2022denoising} like temperature of a softmax function.

%% file: sec/5_experiments.tex
\section{Experiments}

\input{tables/1_main}
\input{tables/2_3dpw_occ}
\input{tables/3_probabilistic}

\input{tables/4_unconditional}
\input{tables/5_ablation}

\subsection{Experimental setup}

\paragraph{Datasets.}
LieHMR is trained in supervised and self-supervised manner, simultaneously. We use the standard datasets, including Human3.6M \cite{h36m_pami}, MPI-INF-3DHP \cite{mono-3dhp2017}, MSCOCO \cite{lin2015microsoftcococommonobjects}, and MPII \cite{andriluka14cvpr} for supervised learning and the large-scale mocap dataset, AMASS \cite{AMASS:ICCV:2019} for self-supervised learning. We uniformly select one dataset from the collection of these datasets and then randomly sample instances from the chosen dataset. We evaluate LieHMR on the in-the-wild 3DPW \cite{vonMarcard2018} and EMDB \cite{kaufmann2023emdb} datasets without finetuning on their training sets. 

\paragraph{Implementation details.}
Following recent works, we use HRNet \cite{sun2019deep} and ViT \cite{dosovitskiy2020vit} as the image backbone. We use a fully connected layer to map the dimension, resulting in $g(I) \in \mathbb{R}^{WH \times d}$ ($W=H=7$ for HRNet and $W=12$, $H=16$ for ViT). For the sequence model, we use transformer-based encoder with 8 blocks and decoder with 8 blocks following ViT \cite{dosovitskiy2020vit} architecture. For the denoising model, we use a small MLP with residual connection. We concatenate the input and a latent vector from the sequence model before every layer and the time embedding is conditioned via AdaLN \cite{Peebles2022DiT}. We train LieHMR in end-to-end manner for about 5 days using 2 NVIDIA RTX 4090 GPUs.

\paragraph{Evaluation metrics.}
We use the Mean Per Joint Position Error (MPJPE), Procrustes-Aligned Mean Per Joint Position Error (PA-MPJPE), and Mean Vertex Error (MVE) in mm to evaluate image-conditioned human pose and shape generation. Procrustes-Alignment is a rigid transformation that minimizes the distance between the prediction and ground-truth. We use the Average Pairwise Distance (APD) for pose diversity and Fréchet distance (FID) for realism to evaluate unconditional human pose and shape generation.

\subsection{Image-conditioned generation}
\label{subsec:5.2}
We compare LieHMR with prior deterministic and probabilistic HMR methods and validate that LieHMR can model well-aligned distribution to 2D observations. Our goal is that LieHMR accurately predicts a single sample from clear images (performance in single-output setting) and various samples from ambiguous images (performance in multiple-output setting). From a fully masked sequence, we predict all tokens at one time, which has $1$ autoregressive step. We use DDIM sampling \cite{song2022denoising} with $1,000$ diffusion timesteps, $\eta=0.0$ for single-output evaluation and $\eta=1.0$ for multiple-output evaluation.

\paragraph{Single-output.}
In \cref{tab:main}, we compare LieHMR with prior deterministic HMR methods including parametric \cite{Kocabas_PARE_2021, goel2023humans, dwivedi_cvpr2024_tokenhmr} and non-parametric \cite{cho2022FastMETRO, Ma_2023_CVPR, li2022cliff, fiche2024vq, fiche2024mega}. We present the results with training on the standard datasets and without finetuning on the 3DPW training split. LieHMR shows slightly worse MVE than non-parametric methods, however the joint position errors are more significant metrics for HMR. With an HRNet backbone, we outperform prior deterministic HMR methods except MEGA \cite{fiche2024mega}, which is a concurrent work. With a ViT backbone, we acheive the state-of-the-art performance in MPJPE. Compared to MEGA, LieHMR exhibits a 1\% reduction on 3DPW and 4\% reduction on EMDB, emphasizing the advantage of parametric methods on generalization.

In \cref{tab:occ}, we also evaluate LieHMR on the 3DPW-OCC \cite{3DPW-OCC} dataset, which is a subset of 3DPW with occlusions. Similar to MEGA, we concatenate the image features to the input sequence and implement self-attention on the concatenated sequence. The transformer learns to determine which joints rely more on the image features or the other joints. Therefore, the attention mechanism enables LieHMR to be robust on the occlusion.

\begin{figure}
  \centering
  \includegraphics[width=1.0\linewidth]{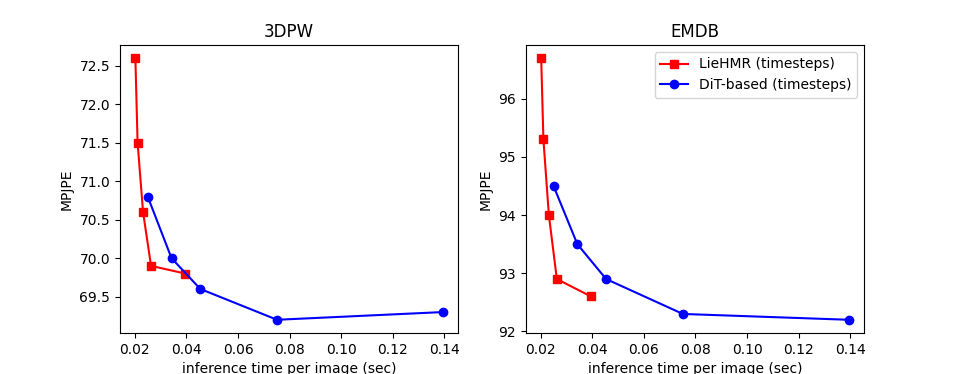}
  \caption{\textbf{Ablation study on image-conditioned generation.} We plot the speed/MPJPE trade off on 10\% subset of 3DPW and EMDB dataset in single-output setting. The curves are obtained by different diffusion timesteps (75, 150, 250, 500, and 1,000).}
  \label{fig:ablation_conditional}
\end{figure}

\begin{figure}
  \centering
  \includegraphics[width=1.0\linewidth]{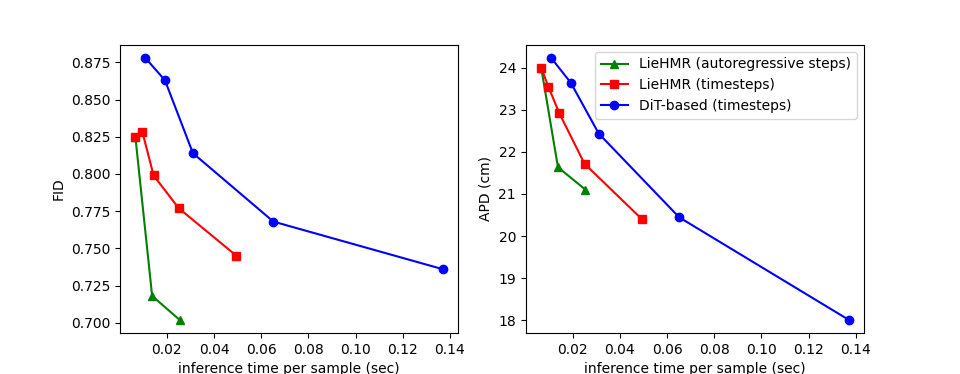}
  \caption{\textbf{Ablation study on unconditional generation.} We plot the speed/FID and speed/APD trade off. The curves are obtained by different diffusion timesteps (75, 150, 250, 500, and 1,000) and autoregressive steps (1, 3, and 6).}
  \label{fig:ablation_unconditional}
\end{figure}

\paragraph{Multiple-output.}
In \cref{tab:multiple}, we compare LieHMR with prior probabilistic HMR methods on the 3DPW dataset. Following prior works \cite{biggs2020multibodies, Cho_2023_ICCV, kolotouros2021prohmr, fiche2024mega}, we report metrics for the best output from $Q$ predictions. LieHMR outperforms normalizing flows \cite{biggs2020multibodies, kolotouros2021prohmr, sengupta2023humaniflow, dwivedi_3dv2023_poco} and diffusion \cite{Cho_2023_ICCV, xu2024scorehypo} based methods. Normalizing flows can compute exact likelihood, however the likelihood does not semantically capture confidence \cite{dwivedi_3dv2023_poco}. Compared to normalizing flows, diffusion based methods have several advantages, such as flexible architecture designs and stable training. Prior diffusion based methods, which are not constrained on $SO(3)$ manifold, are challenging to learn 3D rotation representations. Compared to MEGA, which is autoregressive framework, LieHMR produces more diverse hypotheses, showing higher relative improvement and obtaining better performance from multiple samples because the randomness of MEGA is only related to the sequence order without temperature scaling of softmax function. Therefore, $SO(3)$ diffusion process of LieHMR is the most appropriate to learn the probability distribution of 3D rotation representations. 

\subsection{Unconditional generation}
\begin{figure*}
  \centering
  \includegraphics[width=1.0\linewidth]{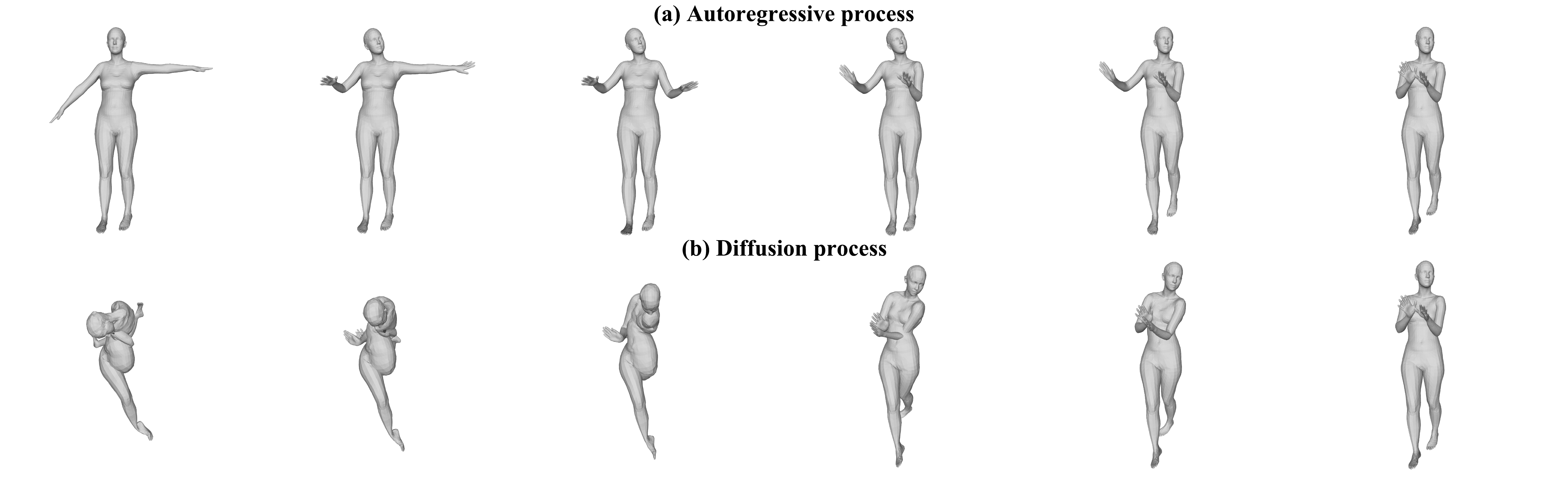}
  \caption{
    \textbf{Generation process.} We visualize the generation process by (a) autoregressive steps and (b) diffusion steps. We replace the masked tokens to identity matrices.
  }
  \label{fig:generation}
\end{figure*}

Following NRDF \cite{he24nrdf}, we measure FID and APD to evaluate generated poses. NRDF shows the state-of-the-art performance with balancing the trade-off between realism (FID) and diversity (APD). For example, VPoser \cite{SMPL-X:2019} shows the lowest FID with less diversity and Pose-NDF \cite{tiwari22posendf} shows the highest APD with less realism. We achieve comparable performance to NRDF while supporting image-conditoned and unconditional generation simultaneously. GFPose-A \cite{ci2022gfpose} is a diffusion-based methods trained on AMASS dataset in joint-location representation (Euclidean space). GFPose-Q is trained in quaternion representation and DPoser \cite{lu2023dposer} is trained in axis-angle representation (non-Euclidean space). LieHMR outperforms GFPose and DPoser in both realism and diversity, which suggests that learning 3D rotation representations with $SO(3)$ diffusion is more suitable for pose priors. Similar to multiple-output image-conditioned generation, LieHMR also generates more diverse samples compared to MEGA. We visualize the generation process in \cref{fig:generation}.

\subsection{Ablation study}

\paragraph{$\boldsymbol{SO(3})$ Diffusion.}
To the best of our knowledge, LieHMR is the first work that is based on $SO(3)$ diffusion model to learn probability distribution of human poses. We compare LieHMR to Euclidean diffusion model with the same architecture. Diffusion models inherently iterate function evaluations, following the reverse process. Therefore, we need to project the intermediate data to proper manifold and design the forward and reverse process to be constrained on the manifold. We validate the advantage of $SO(3)$ diffusion model on image-conditioned generation in \cref{tab:ablation_so3}

\paragraph{Disentangling the transformer and denoising model.}
The computational complexity of transformer-based diffusion models is $O(N^2T)$, where $N$ is the sequence length and $T$ is the number of diffusion timesteps. These models learn the joint distribution of the entire sequence, making the transformer time-dependent and requiring the forward pass to be repeated $T$ times. However, we disentangle the transformer and denoising model, reducing the complexity to $O(N^2)+O(T)$. We demonstrate the speed/accuracy trade-off at inference time between ours and DiT \cite{Peebles2022DiT} based design (in \cref{fig:models} left) with varying diffusion timesteps. First, we compare the performance of image-conditioned generation in \cref{fig:ablation_conditional}. Our method provides more accurate results under inference budgets, especially on more challenging EMDB dataset. Second, we also compare the unconditional generation in \cref{fig:ablation_unconditional} and our method generates much more realistic samples with similar diversity under inference budgets. In addition to the efficiency, learning the complex joint distribution, which is product of $SO(3)$ manifolds, is much more challenging than learning the per-joint distribution.

\paragraph{Comparing to Vector-Quantization based methods.}
Similar to Vector-Quantization based methods (in \cref{fig:models} right), LieHMR adopts the transformer to model the probability distribution of each token. However, Vector-Quantization based methods learn categorical distributions using VQ-VAE, and LieHMR learns the per-joint distribution using $SO(3)$ diffusion. While, VQ-HPS \cite{fiche2024vq} and MEGA \cite{fiche2024mega} predicts 54 tokens among 512 values, TokenHMR \cite{dwivedi_cvpr2024_tokenhmr} 160 tokens among 2048 values, and GenHMR \cite{saleem2024genhmrgenerativehumanmesh} 96 tokens among 2048, LieHMR extracts 24 latent vectors with size 512 to model probability distribution of 24 tokens. We achieve powerful performance compared to Vector-Quantization based methods with fewer tokens.

\paragraph{Self-supervised training.}
Several works \cite{wei2023diffusion, xie2024showo, wu2024vila, wu2024janus, ma2024janusflow, chen2025janus} have demonstrated that generation can facilitate understanding. Inspired by recent HMR methods \cite{dwivedi_cvpr2024_tokenhmr, fiche2024vq, fiche2024mega, saleem2024genhmrgenerativehumanmesh}, we also adopt the mocap dataset, AMASS to learn pose priors by unconditional generation. In addition to unconditional generation, the self-supervised training slightly improves the image-conditioned generation as shown in \cref{tab:ablation_amass}. The learned pose priors constrain the valid space during training and inference.

\paragraph{Autoregressive steps.}
We can balance speed and accuracy at inference time by controlling the number of autoregressive steps. For the image-conditioned generation, more iterations in autoregressive process slightly improve the performance with more samples, as shown in \cref{tab:ablation_occ}. The attention mechanism of LieHMR is self-attention on the concatenated sequence of pose tokens and image features. We assume that there are more attention weights on the image features given clear images and on the visible joints given ambiguous images. On the other hand, more iterations generate more realistic but less diverse samples for the unconditional generation in \cref{fig:ablation_unconditional}. Conditioned on the visible tokens, the autoregressive progress improves the realism and decrease the diversity.

%% file: tables/1_main.tex
\begin{table*}[]
\centering
\begin{tabular}{ll|ccc|ccc}
\toprule
\multicolumn{2}{l|}{}     & \multicolumn{3}{c|}{3DPW} & \multicolumn{3}{c}{EMDB} \\
Method         & Backbone & MPJPE $\downarrow$ & PA-MPJPE $\downarrow$ & MVE $\downarrow$ & MPJPE $\downarrow$ & PA-MPJPE $\downarrow$ & MVE $\downarrow$ \\
\midrule

FastMETRO-L \cite{cho2022FastMETRO}
& HRNet     & 109.0 & 65.7 & 121.6     & 108.1 & 66.8 & 119.2     \\

PARE \cite{Kocabas_PARE_2021}
& HRNet      & 82.0  & 50.9 & 97.9     & 113.9 & 72.2 & 133.2     \\

Virtual Marker \cite{Ma_2023_CVPR}
& HRNet      & 80.5  & 48.9 & 93.8     & -     & -    & -        \\

CLIFF \cite{li2022cliff}
& HRNet      & 73.9  & 46.4 & 87.6     & 103.1 & 68.8 & 122.9     \\

VQ-HPS \cite{fiche2024vq}
& HRNet      & 72.2  & 45.2 & \underline{84.8}     & 99.9  & 65.2 & \underline{112.9}     \\

MEGA$^\dagger$ \cite{fiche2024mega}
& HRNet      & \textbf{68.5}  & \textbf{44.1} & \textbf{81.6}     & \textbf{90.5}  & \textbf{58.7} & \textbf{107.9}     \\

LieHMR (ours)
& HRNet      & \underline{70.1}  & \underline{44.5} & 90.9     & \underline{93.7}  & \underline{59.8} & 121.0     \\
\midrule

HMR2.0 \cite{goel2023humans}
& ViT        & 70.0  & 44.5 & \underline{84.1}     & 97.8  & 61.5 & 120.1     \\

TokenHMR$^\ddagger$ \cite{dwivedi_cvpr2024_tokenhmr}
& ViT        & 71.0  & 44.3 & 84.6     & \underline{91.7}  & 55.6 & \underline{109.4}     \\

MEGA$^\dagger$ \cite{fiche2024mega}
& ViT        & \underline{67.5}  & \textbf{41.0} & \textbf{80.0}     & 92.4  & \textbf{52.5} & \textbf{108.6}     \\

LieHMR (ours)
& ViT        & \textbf{66.9}  & \underline{42.6} & 86.6     & \textbf{88.4} & \underline{55.4}  & 114.0 \\
\bottomrule
\end{tabular}
\caption{\textbf{Evaluation in single-output setting.} We evaluate LieHMR on 3DPW and EMDB datasets compared to SOTA deterministic methods. $\dagger$ stands for concurrent work and $\ddagger$ for using additional training data. Best in bold, second-best underlined.}
\label{tab:main}
\end{table*}

%% file: tables/2_3dpw_occ.tex
\begin{table}[]
\resizebox{\columnwidth}{!}{
\begin{tabular}{l|ccc}
\toprule
Method       & MPJPE $\downarrow$ & PA-MPJPE $\downarrow$ & MVE $\downarrow$ \\
\midrule

ROMP \cite{ROMP}
& -     & 65.9  & -        \\

SPIN \cite{kolotouros2019spin}
& 95.6  & 60.8 & 121.6     \\

VisDB \cite{yao2022learning}
& 87.3  & 56.0 & 110.5     \\

3DCrowdNet \cite{choi2022learning}
& 88.6  & 56.8 & 103.2     \\

SEFD \cite{SEFD}
& 83.5  & 55.0  & 97.1     \\

MEGA$^\dagger$ \cite{fiche2024mega}
& 79.8  & 51.5  & 93.8     \\
\midrule

PARE \cite{Kocabas_PARE_2021}
& 90.5  & 57.1 & 107.9     \\

ScoreHypo \cite{xu2024scorehypo}
& 73.9  & 48.7  & 89.8     \\

MEGA$^\dagger$ \cite{fiche2024mega}
& \textbf{66.3}  & \underline{43.7}  & \textbf{78.9}     \\

LieHMR (ours)
& \underline{67.4}  & \textbf{42.7}  & \underline{88.2}     \\
\bottomrule
\end{tabular}
}
\caption{\textbf{Evaluation on 3DPW-OCC.} We evaluate LieHMR on 3DPW-OCC dataset, which contains challenging occluded samples. Methods above the line use ResNet backbone, and the others use HRNet. $\dagger$ stands for concurrent work. Best in bold, second-best underlined.}
\label{tab:occ}
\end{table}

%% file: tables/3_probabilistic.tex
\begin{table*}[]
\begin{adjustbox}{scale=0.96}
\begin{tabular}{ll|cccc|cccc|cccc}
\toprule
\multicolumn{2}{c|}{$Q$}  & 1     & 5     & 10    & 25    & 1     & 5    & 10   & 25   & 1     & 5     & 10    & 25   \\
\midrule
Method         & Backbone & \multicolumn{4}{c|}{MPJPE $\downarrow$} & \multicolumn{4}{c|}{PA-MPJPE $\downarrow$} & \multicolumn{4}{c}{MVE $\downarrow$} \\
\midrule

Diff-HMR \cite{Cho_2023_ICCV}
& ResNet   & 98.9  & 96.3 & 95.5 & 94.5 & 58.5  & 57.0  & 56.5  & 55.9 & 114.6 & 111.8 & 110.9 & 109.8 \\

3D Multibodies \cite{biggs2020multibodies}
& ResNet   & 93.8  & 82.2 & 79.4 & 75.8 & 59.9  & 57.1  & 56.6  & 55.6 & -     & -     & -     & -     \\

ProHMR \cite{kolotouros2021prohmr}
& ResNet   & 97.0  & 93.1 & 89.8 & 84.0 & 59.8  & 56.5  & 54.6  & 52.4 & -     & -     & -     & -     \\

HuManiFlow \cite{sengupta2023humaniflow}
& ResNet   & 83.9  & -    & -    & -    & 53.4  & -     & -     & -    & -     & -     & -     & -     \\

POCO \cite{dwivedi_3dv2023_poco}
& ResNet   & 88.5  & -    & -    & -    & 52.4  & -     & -     & -    & 101.1 & -     & -     & -     \\

MEGA$^\dagger$ \cite{fiche2024mega}
& ResNet   & 86.2  & 78.0 & 76.4 & 73.9 & 58.6  & 51.6  & 49.7  & 47.6 & 101.6 & 92.8  & 90.4  & 87.5  \\
\midrule

POCO \cite{dwivedi_3dv2023_poco}
& HRNet   & 80.3  & -    & -    & -    & 49.9  & -     & -     & -    & 95.3  & -     & -     & -     \\

ScoreHypo \cite{xu2024scorehypo}
& HRNet    & -     & 75.3 & 71.7 & 67.8 & -     & 47.4  & 45.2  & 42.5 & -     & -     & -     & -     \\

MEGA$^\dagger$ \cite{fiche2024mega}
& HRNet    & \textbf{69.9}  & \textbf{66.1} & \underline{64.9} & \underline{63.6} & \textbf{45.5}  & \underline{42.6}  & \underline{41.7}  & \textbf{40.4} & \textbf{83.4}  & \textbf{78.4}  & \textbf{76.9}  & \textbf{75.1}  \\

LieHMR (ours)
& HRNet    & \underline{72.0}  & \underline{66.2} & \textbf{64.5} & \textbf{62.6} & \underline{45.8}  & \textbf{42.5}  & \textbf{41.5}  & \textbf{40.4} & \underline{93.5}  & \underline{86.3}  & \underline{84.1}  & \underline{81.7}  \\
\bottomrule
\end{tabular}
\end{adjustbox}
\caption{\textbf{Evaluation in multiple-output setting.} We evaluate LieHMR on 3DPW dataset compared to SOTA probabilistic methods. $\dagger$ stands for concurrent work. Best in bold, second-best underlined.}
\label{tab:multiple}
\end{table*}

%% file: tables/4_unconditional.tex
\begin{table}[]
\centering
\begin{tabular}{l|cc}
\toprule
Method   & FID $\downarrow$  & APD $\uparrow$ (in cm) \\ 
\midrule

GMM \cite{bogo2016smplautomaticestimation3d}
& 0.435 $^{\pm.017}$ & 21.944 $^{\pm.102}$  \\

VPoser \cite{SMPL-X:2019}
& 0.048 $^{\pm.002}$ & 14.684 $^{\pm.138}$  \\

GAN-S \cite{davydov2021adversarialparametricposeprior}
& 0.201 $^{\pm.030}$ & 10.914 $^{\pm.396}$  \\

Pose-NDF \cite{tiwari22posendf}
& 3.920 $^{\pm.034}$ & 37.813 $^{\pm.085}$  \\

GFPose-A \cite{ci2022gfpose}
& 1.246 $^{\pm.005}$ & 13.876 $^{\pm.116}$  \\

GFPose-Q \cite{ci2022gfpose}
& 1.635 $^{\pm.002}$ & 6.773 $^{\pm.112}$   \\

FM-Dis \cite{chen2023riemannianfm}
& 0.346 $^{\pm.007}$ & 6.849 $^{\pm.199}$   \\

NRDF \cite{he24nrdf}
& 0.636 $^{\pm.007}$ & 23.116 $^{\pm.105}$  \\

DPoser \cite{lu2023dposer}
& -     & 19.03    \\

MEGA \cite{fiche2024mega}
& -     & 20.77    \\

LieHMR (ours)
& 0.825 $^{\pm.067}$ & 23.999 $^{\pm.602}$ \\
\bottomrule
\end{tabular}
\caption{\textbf{Unconditional generation.} We sample $20 \times 500$ Human Pose and Shape. Our performance is obtained from $1$ autoregressive step, $75$ diffusion timesteps and $\eta=1.0$. $\pm$ indicates the $95\%$ confidence interval.}
\label{tab:unconditional}
\end{table}

%% file: tables/5_ablation.tex
\begin{table}[]
\centering
% \begin{adjustbox}{scale=1.0}
\begin{tabular}{l|ccc}
\toprule
& MPJPE $\downarrow$ & PA-MPJPE $\downarrow$ & MVE $\downarrow$ \\
\midrule

LieHMR (ours) & \textbf{92.6} & \textbf{59.1} & \textbf{119.6} \\
w/o AMASS     & \textbf{92.6} & 59.5 & 120.0 \\

\bottomrule
\end{tabular}
% \end{adjustbox}
\caption{\textbf{Ablation study of self-supervised training.} We report the evaluation metrics on 10\% subset of EMDB in single-output setting. Best in bold.}
\label{tab:ablation_amass}
\end{table}

\begin{table}[]
\centering
\begin{tabular}{l|cccc}
\toprule
steps & $Q=1$    & $Q=5$    & $Q=10$   & $Q=25$   \\
\midrule

1     & \textbf{71.7} & 66.0 & 64.2 & 62.3 \\
3     & \textbf{71.7} & \textbf{65.7} & 64.0 & 62.1 \\
6     & 71.8 & \textbf{65.7} & \textbf{63.9} & \textbf{61.8} \\
\bottomrule
\end{tabular}
\caption{\textbf{Ablation study of autoregressive steps.} We report the MPJPE on 10\% subset of 3DPW in multiple-output setting with different autoregressive steps. Best in bold.}
\label{tab:ablation_occ}
\end{table}

\begin{table}[]
\begin{adjustbox}{width=\linewidth}
\centering
\begin{tabular}{l|cc|cc}
\toprule
 & \multicolumn{2}{c|}{3DPW} & \multicolumn{2}{c}{EMDB} \\
Method & MPJPE $\downarrow$ & PA-MPJPE $\downarrow$ & MPJPE $\downarrow$ & PA-MPJPE $\downarrow$ \\
\midrule

Euclidean diffusion & 74.5 & 44.8 & 103.9 & 60.1 \\
$SO(3)$ diffusion & \textbf{70.1} & \textbf{44.5} & \textbf{93.7} & \textbf{59.8} \\
\bottomrule

\end{tabular}
\end{adjustbox}
\caption{\textbf{Ablation study on diffusion.} We compare $SO(3)$ diffusion to Euclidean diffusion on image-conditioned generation. Best in bold.}
\label{tab:ablation_so3}
\end{table}

%% file: sec/6_conclusion.tex
\section{Conclusion}

In this paper, we propose LieHMR, which generates the well-aligned distribution of human pose and shape parameters to 2D observations. LieHMR can produce multiple plausible outputs given a highly ambiguous image and accurate single output given a clear image. LieHMR is based on $SO(3)$ diffusion to learn image-conditioned and unconditional distribution via conditioning dropout. We disentangle the transformer and denoising model for better speed/accuracy trade-off. The extensive experiments demonstrate that LieHMR has strong advantages on single-output/multiple-output HMR and unconditional pose generation.

\paragraph{Limitations and future work.}
The diffusion process is computationally expensive at inference time although DDIM \cite{song2022denoising} with smaller diffusion timesteps enables more efficient sampling. We plan to adopt Flow Matching \cite{lipman2023flow, chen2023riemannianfm} on LieHMR. Recently, autoregressive models have emerged as universal multimodal models, capable of processing images, videos, text, audio, and more \cite{videopoet, lyu2023macaw}. Our future research could extend LieHMR into a universal motion generation model conditioned on images, videos, text, and audio \cite{zhou2022udeunifieddrivingengine, anisetty2024dynamicmotionsynthesismasked, yang2024unimumounifiedtextmusic}.

\section{Acknowledgements}
This work was supported by NST grant (CRC 21011, MSIT), IITP grant (RS-2023-00228996, RS-2024-00459749, RS-2025-25443318, RS-2025-25441313, MSIT) and KOCCA grant (RS-2024-00442308, MCST).

%% file: sec/X_suppl.tex
\clearpage
\setcounter{page}{1}
\maketitlesupplementary

\setcounter{figure}{0}
\setcounter{table}{0}
\counterwithin{figure}{section}
\counterwithin{table}{section}

\renewcommand{\thesection}{S}
\renewcommand{\thetable}{S\arabic{table}}
\renewcommand{\thefigure}{S\arabic{figure}}

\begin{figure}
  \centering
  \includegraphics[width=0.85\linewidth]{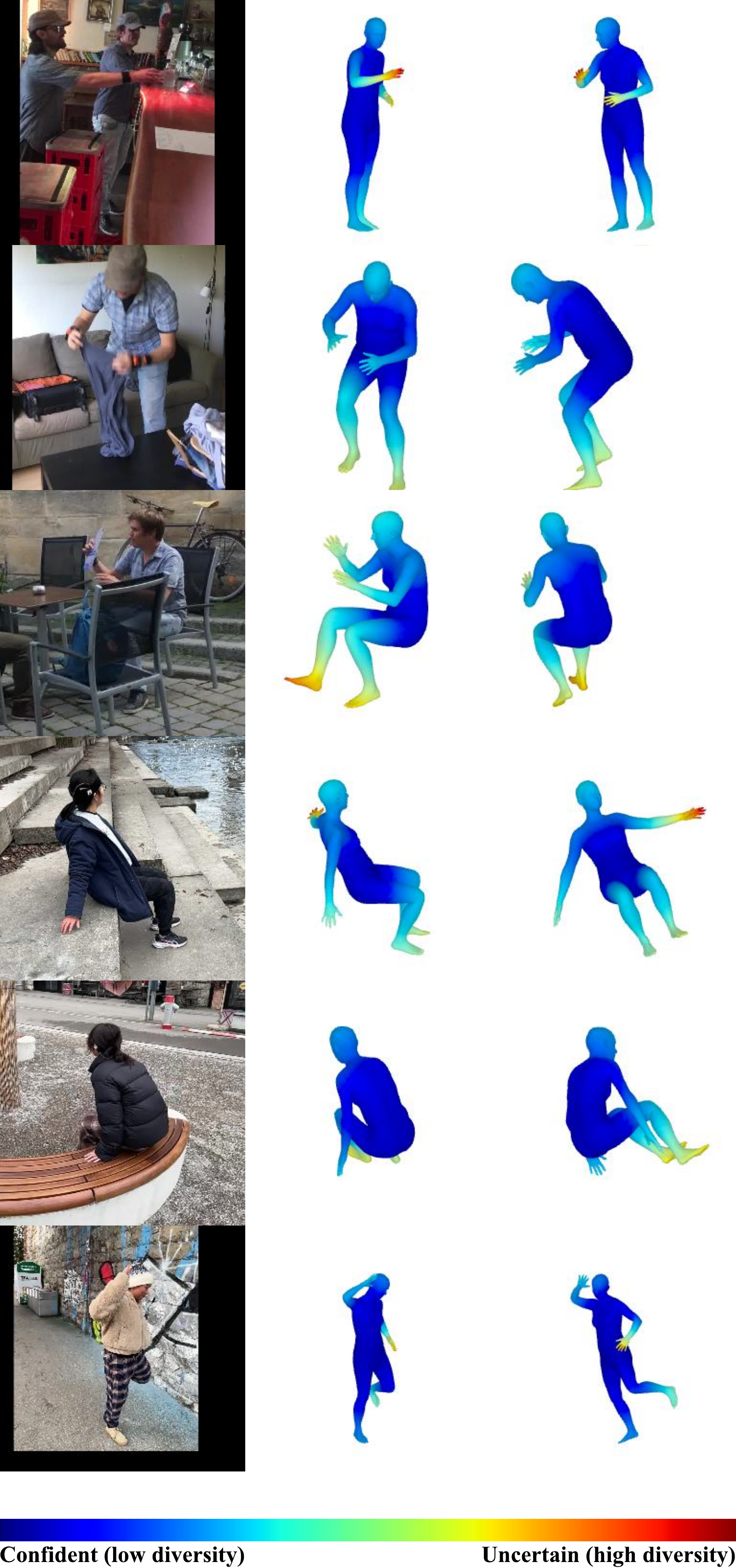}
  \caption{\textbf{Visualization of the uncertainty.}}
  \label{fig:vertex_std}
\end{figure}

\subsection{Diversity in predictions}
Following the multiple-output setting in \cref{subsec:5.2}, we generate 25 samples from given images. We compute the standard deviation of 3D position of each vertex and visualize the values on the meshes in \cref{fig:vertex_std}. We normalize the standard deviation per each batch. The standard deviation can be considered as epistemic uncertainty \cite{uncertainty}. The body parts with occlusion or depth ambiguity show higher diversity.

\begin{figure}
  \centering
  \includegraphics[width=1.0\linewidth]{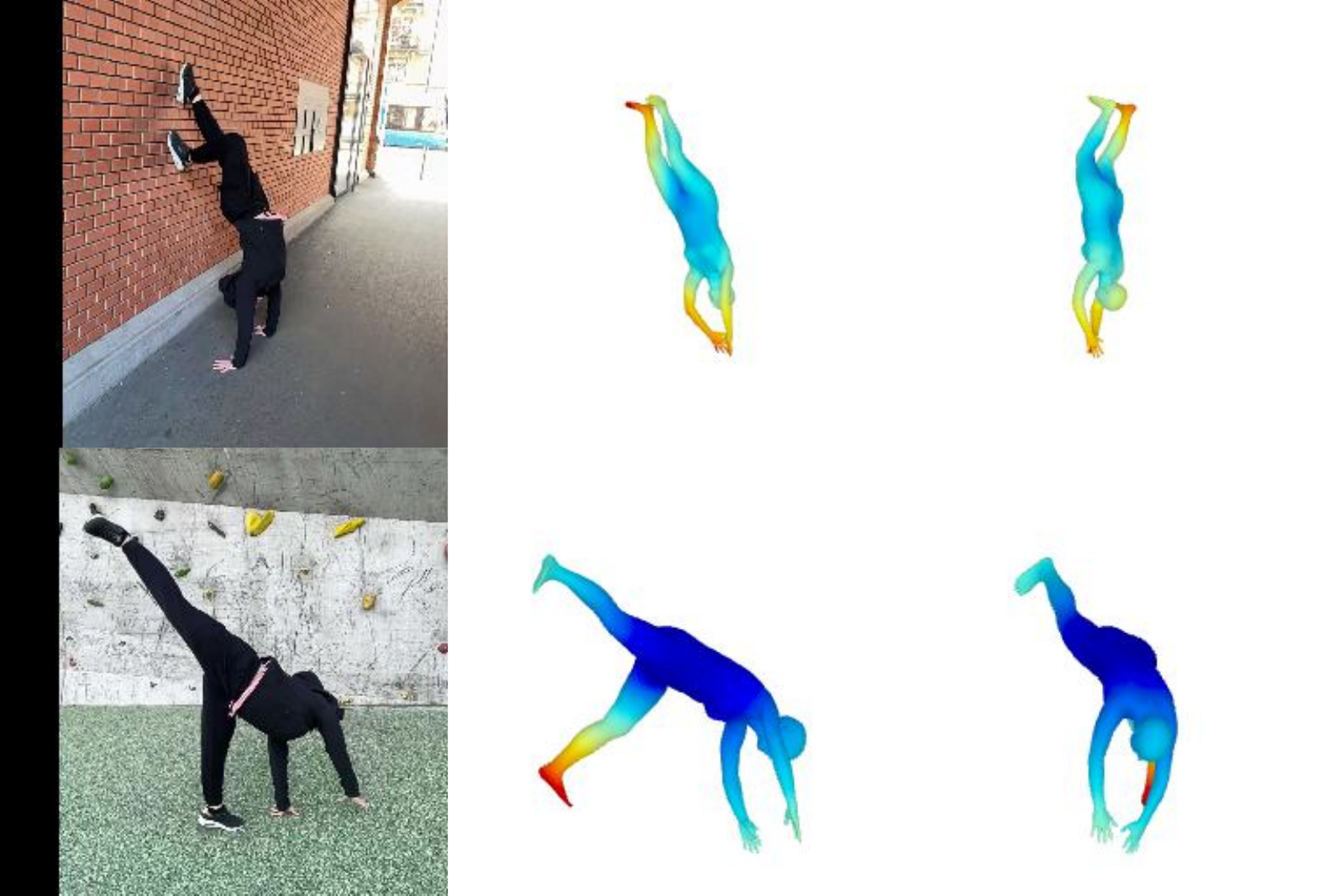}
  \caption{
    \textbf{Failure cases of image-conditioned generation.} 
  }
  \label{fig:failre_case_image}
\end{figure}

\begin{figure}
  \centering
  \includegraphics[width=1.0\linewidth]{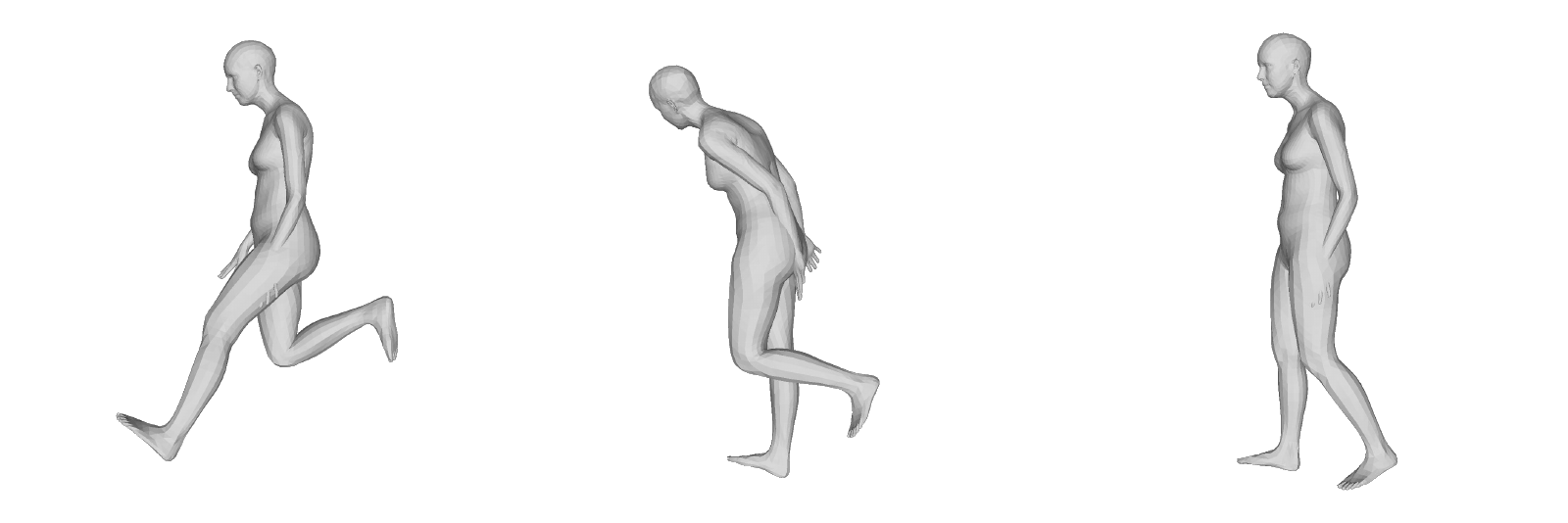}
  \caption{
    \textbf{Failure cases of unconditional generation.} 
  }
  \label{fig:failure_case_unconditional}
\end{figure}

\subsection{Failure cases}
We present several failure cases of image-conditioned generation in \cref{fig:failre_case_image} and unconditional generation in \cref{fig:failure_case_unconditional}. For, image-conditioned generation, LieHMR struggles to produce accurate results for extreme poses. We assume that these are "unseen" poses during training. Training LieHMR on additional datasets with extreme poses (\textit{e.g.}, MOYO \cite{tripathi2023ipman}) could help mitigate this issue. Notably, the uncertainty is high enough to indicate that our model fails. For unconditional generation, LieHMR sometimes generates human meshes with self-penetration. We will try physical simulation or test-time guidance to generate physically plausible samples in our future work.

\begin{figure*}
  \centering
  \includegraphics[width=0.7\linewidth]{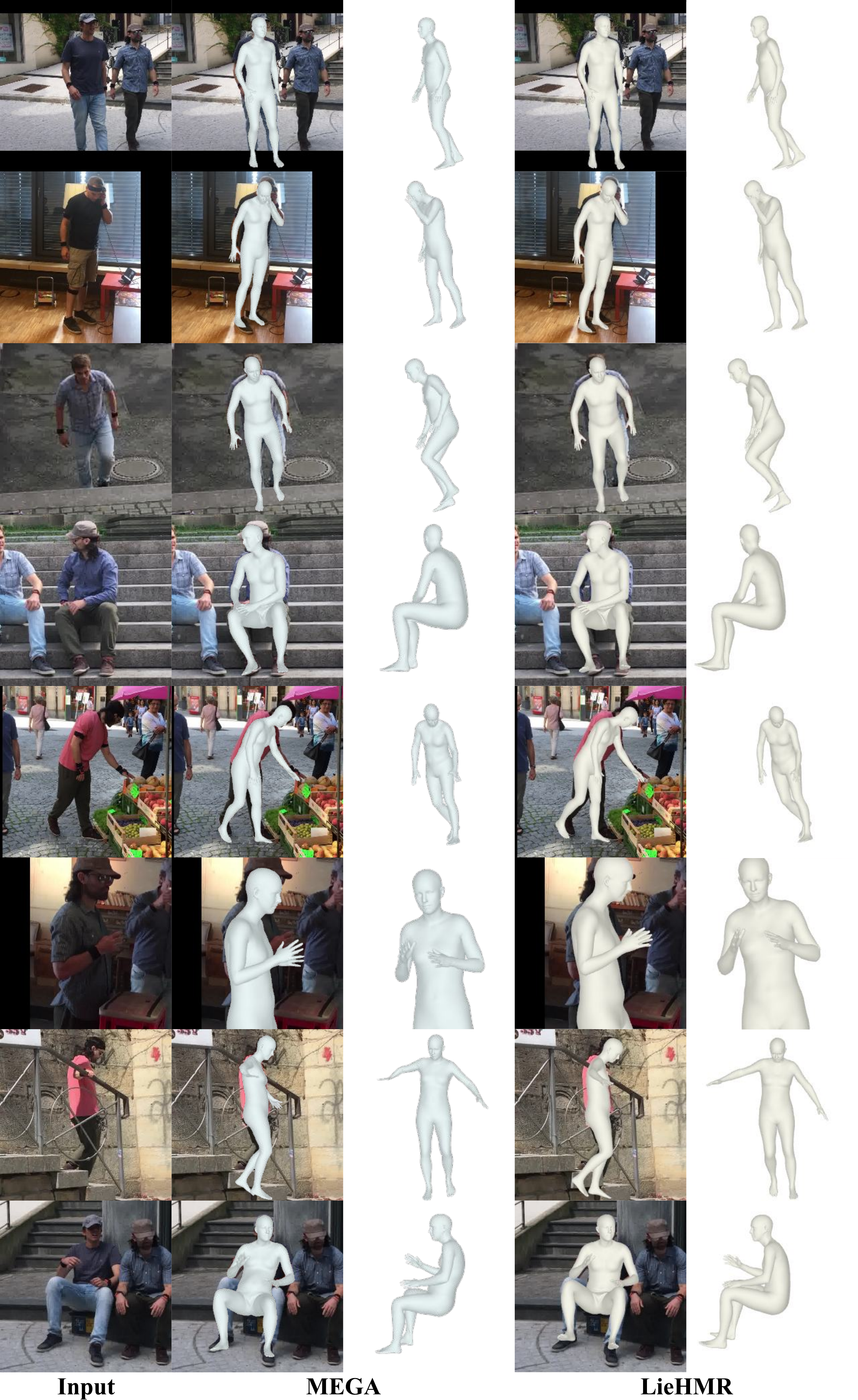}
  \caption{
    \textbf{Qualitative results in single-output setting on 3DPW dataset \cite{vonMarcard2018}.}
  }
  \label{fig:single_output1}
\end{figure*}

\begin{figure*}
  \centering
  \includegraphics[width=0.7\linewidth]{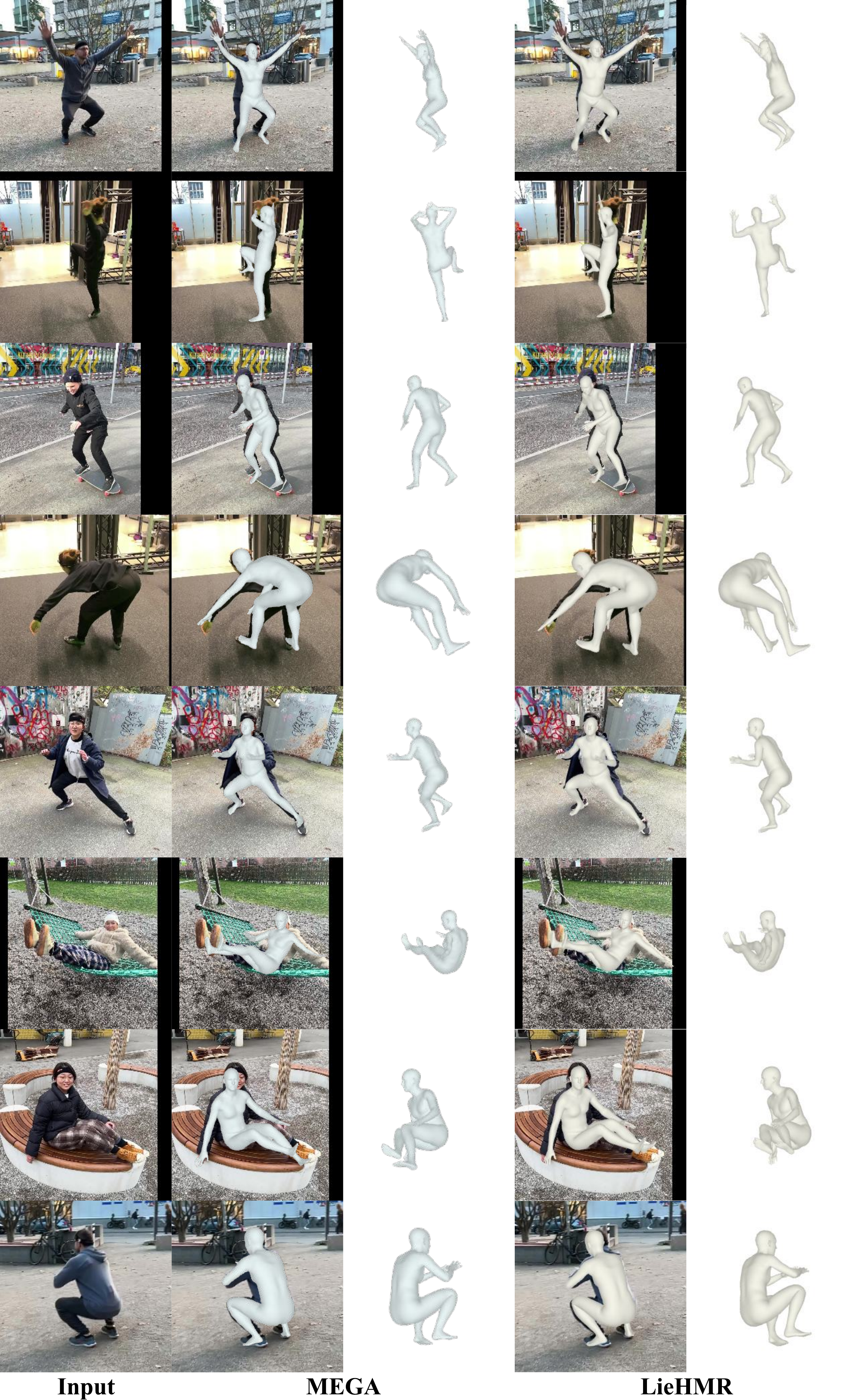}
  \caption{
    \textbf{Qualitative results in single-output setting on EMDB dataset \cite{kaufmann2023emdb}.}
  }
  \label{fig:single_output2}
\end{figure*}

\begin{figure*}
  \centering
  \includegraphics[width=0.7\linewidth]{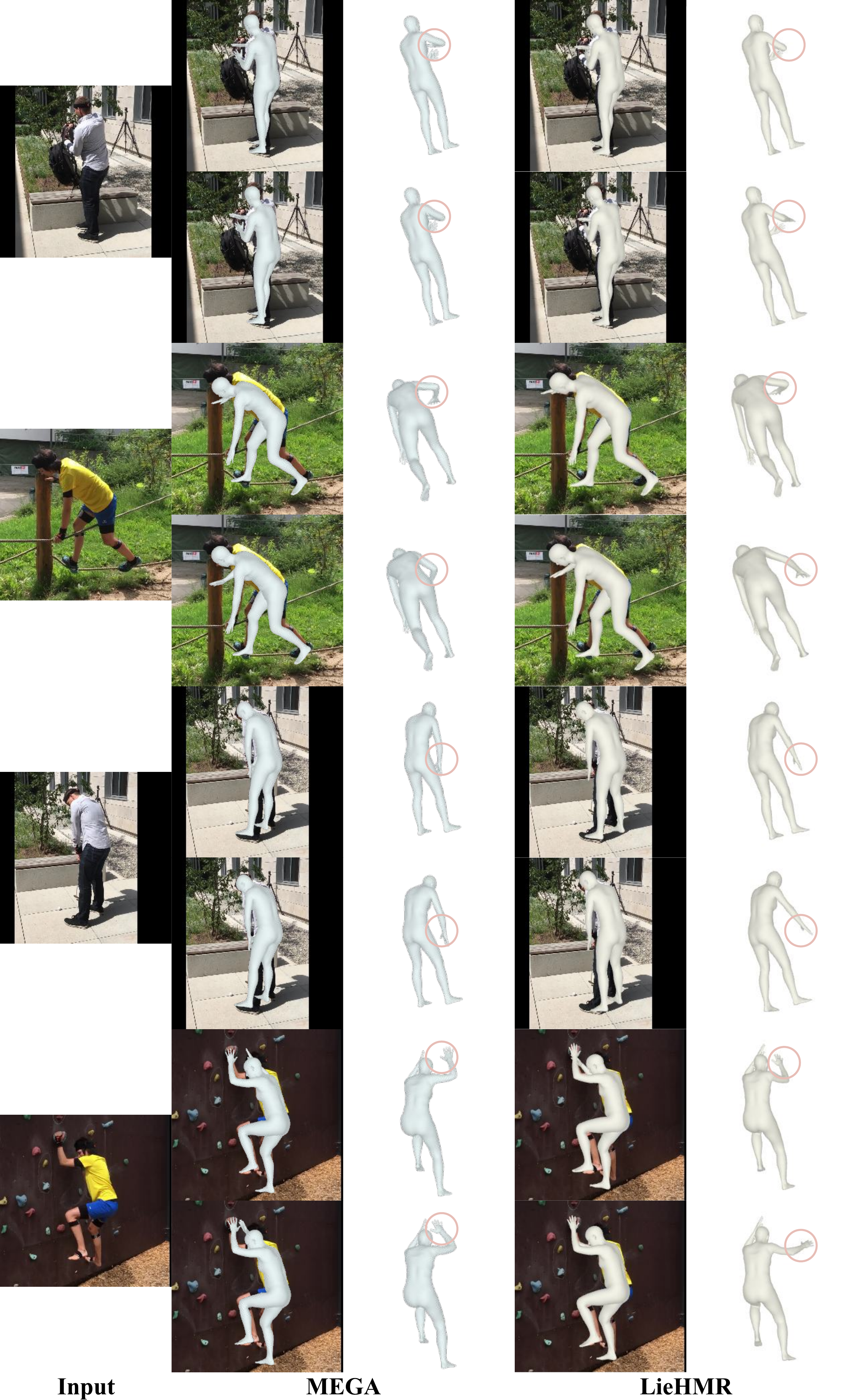}
  \caption{
    \textbf{Qualitative results in multiple-output setting on 3DPW dataset \cite{vonMarcard2018}.}
  }
  \label{fig:multiple_output1}
\end{figure*}

\begin{figure*}
  \centering
  \includegraphics[width=0.7\linewidth]{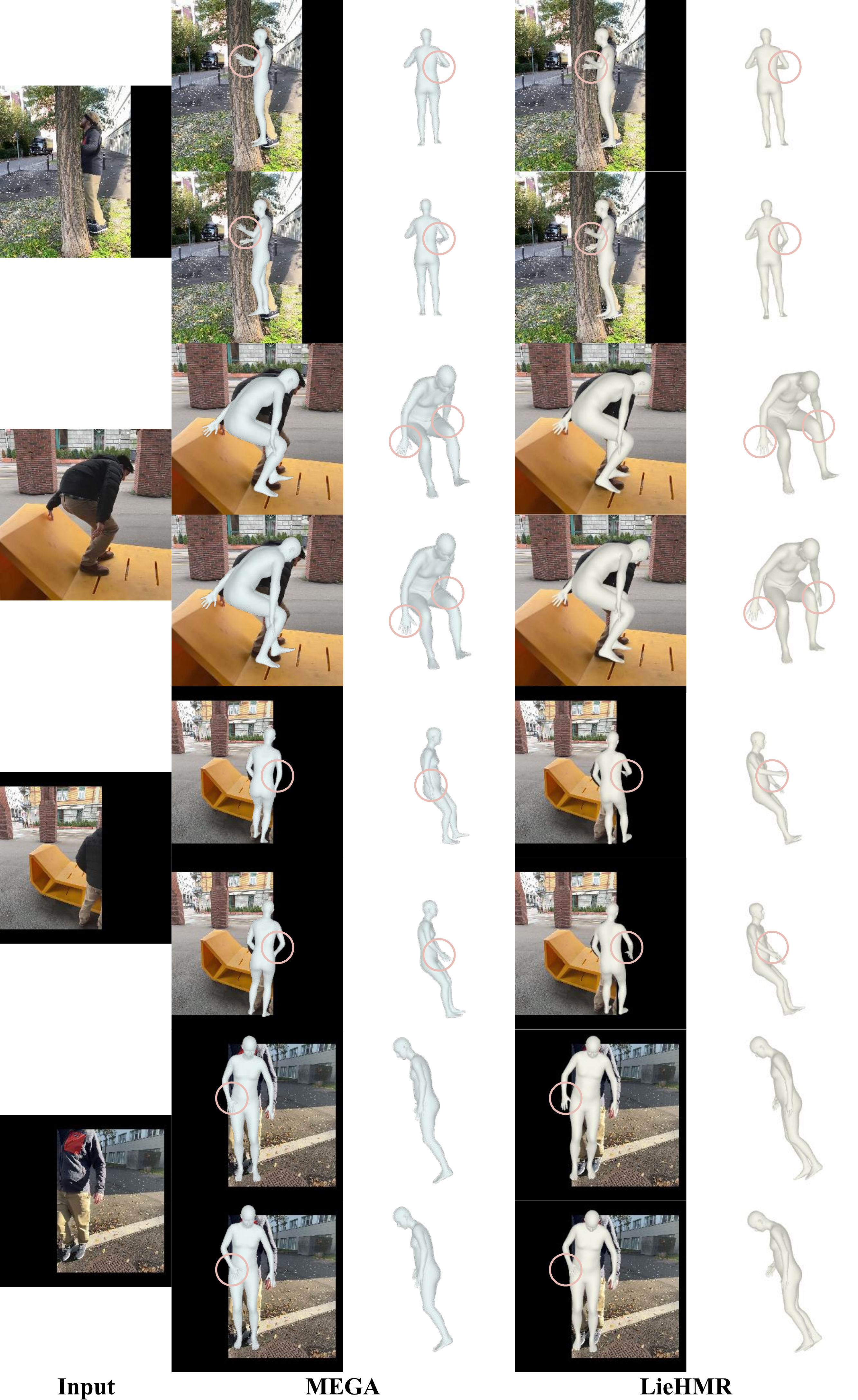}
  \caption{
    \textbf{Qualitative results in multiple-output setting on EMDB dataset \cite{kaufmann2023emdb}.}
  }
  \label{fig:multiple_output2}
\end{figure*}

% \subsection{Qualitative results}
\paragraph{Qualitative results.} We present qualitative results in single-output (in Fig. \ref{fig:single_output1} and \ref{fig:single_output2}) and multiple-output setting (in Fig. \ref{fig:multiple_output1} and \ref{fig:multiple_output2}) compared to MEGA \cite{fiche2024mega}. Similar to quantitative evaluation, LieHMR demonstrates comparable performance to MEGA, while generating much more diverse and plausible samples. As highlighted in Fig. \ref{fig:multiple_output1} and \ref{fig:multiple_output2}, the diversity of generated samples comes from occlusion or depth ambiguity.

